\documentclass{article}
\usepackage{array}
\usepackage{calc}
\newcolumntype{C}[1]{>{\centering\arraybackslash}p{#1}}

\usepackage[letterpaper,
            textheight=9in, textwidth=5.5in,
            top=1in, headheight=12pt, headsep=25pt, footskip=30pt]{geometry}

\usepackage[square,numbers]{natbib}
\usepackage{xcolor}
\usepackage{hyperref}
\definecolor{mydarkblue}{rgb}{0,0.08,0.45}
\hypersetup{
  colorlinks=true,
  linkcolor=mydarkblue, citecolor=mydarkblue,
  filecolor=mydarkblue, urlcolor=mydarkblue,
  pdfborder={0 0 0}, pdfview=FitH
}

\providecommand{\doi}[1]{%
  \begingroup\urlstyle{rm}%
  \href{http://dx.doi.org/#1}{doi:\discretionary{}{}{}\nolinkurl{#1}}%
  \endgroup}

\usepackage[utf8]{inputenc}
\usepackage[T1]{fontenc}
\usepackage[english]{babel}
\usepackage{hyperref}
\usepackage{url}
\usepackage{amsfonts}
\usepackage{nicefrac}
\usepackage{microtype}
\usepackage{amsmath}
\usepackage{amssymb}
\usepackage{mathtools}
\usepackage{subcaption}
\usepackage{booktabs}
\usepackage{ragged2e}
\usepackage[table,dvipsnames]{xcolor}
\usepackage{etoolbox}
\usepackage{xspace}
\usepackage{xpatch}
\usepackage{enumerate}
\usepackage{xstring}
\usepackage{setspace}
\usepackage{tabularx}
\usepackage{graphicx}
\usepackage{stackengine}
\usepackage{pgffor}
\usepackage[capitalise,noabbrev,nameinlink]{cleveref}
\usepackage[useregional=numeric]{datetime2}
\usepackage{wrapfig}
\usepackage[export]{adjustbox}
\usepackage{ifthen}
\usepackage{pifont}
\usepackage{titlesec}
\usepackage[normalem]{ulem}

\usepackage{algorithm}
\usepackage{algpseudocode}

\usepackage{tabularx}
\usepackage{booktabs}
\usepackage{arydshln}
\renewcommand{\arraystretch}{1.2}
\usepackage{booktabs}
\usepackage{multirow}
\usepackage{array}

\usepackage{array}
\newcolumntype{Y}[1]{>{\centering\arraybackslash}p{#1}}
\usepackage{makecell}

\usepackage{tabularx}
\usepackage{booktabs}
\newcommand{\methodname}{CAMP\xspace}
\usepackage{wrapfig}
\usepackage{rotating}
\usepackage{enumitem}
\definecolor{myorange}{rgb}{0.9, 0.45, 0.0} 
\usepackage[cal=cm]{mathalfa}

\makeatletter
\renewcommand{\normalsize}{%
  \@setfontsize\normalsize\@xpt\@xipt
  \abovedisplayskip      7\p@ \@plus 2\p@ \@minus 5\p@
  \abovedisplayshortskip \z@ \@plus 3\p@
  \belowdisplayskip      \abovedisplayskip
  \belowdisplayshortskip 4\p@ \@plus 3\p@ \@minus 3\p@}
\renewcommand{\small}{%
  \@setfontsize\small\@ixpt\@xpt
  \abovedisplayskip      6\p@ \@plus 1.5\p@ \@minus 4\p@
  \abovedisplayshortskip \z@  \@plus 2\p@
  \belowdisplayskip      \abovedisplayskip
  \belowdisplayshortskip 3\p@ \@plus 2\p@ \@minus 2\p@}
\renewcommand{\footnotesize}{\@setfontsize\footnotesize\@ixpt\@xpt}
\renewcommand{\scriptsize}{\@setfontsize\scriptsize\@viipt\@viiipt}
\renewcommand{\tiny}{\@setfontsize\tiny\@vipt\@viipt}
\renewcommand{\large}{\@setfontsize\large\@xiipt{14}}
\renewcommand{\Large}{\@setfontsize\Large\@xivpt{16}}
\renewcommand{\LARGE}{\@setfontsize\LARGE\@xviipt{20}}
\renewcommand{\huge}{\@setfontsize\huge\@xxpt{23}}
\renewcommand{\Huge}{\@setfontsize\Huge\@xxvpt{28}}
\normalsize

\renewcommand{\section}{\@startsection{section}{1}{\z@}%
  {-2.0ex \@plus -0.5ex \@minus -0.2ex}%
  { 1.5ex \@plus  0.3ex \@minus  0.2ex}%
  {\large\bf\raggedright}}
\renewcommand{\subsection}{\@startsection{subsection}{2}{\z@}%
  {-1.8ex \@plus -0.5ex \@minus -0.2ex}%
  { 0.8ex \@plus  0.2ex}%
  {\normalsize\bf\raggedright}}
\renewcommand{\subsubsection}{\@startsection{subsubsection}{3}{\z@}%
  {-1.5ex \@plus -0.5ex \@minus -0.2ex}%
  { 0.5ex \@plus  0.2ex}%
  {\normalsize\bf\raggedright}}
\renewcommand{\paragraph}{\@startsection{paragraph}{4}{\z@}%
  {1.5ex \@plus 0.5ex \@minus 0.2ex}{-1em}{\normalsize\bf}}

\renewenvironment{table}
  {\setlength{\abovecaptionskip}{\z@}%
   \setlength{\belowcaptionskip}{7\p@}%
   \@float{table}}
  {\end@float}

\renewcommand{\@maketitle}{%
  \vbox{\hsize\textwidth \linewidth\hsize \vskip 0.1in \centering
    {\LARGE\bf \@title\par}
    \def\And{\end{tabular}\hfil\linebreak[0]\hfil%
      \begin{tabular}[t]{c}\bf\rule{\z@}{24\p@}\ignorespaces}
    \def\AND{\end{tabular}\hfil\linebreak[4]\hfil%
      \begin{tabular}[t]{c}\bf\rule{\z@}{24\p@}\ignorespaces}
    \begin{tabular}[t]{c}\bf\rule{\z@}{24\p@}\@author\end{tabular}%
    \vskip 0.3in \@minus 0.1in}}

\renewcommand{\@maketitle}{%
  \vbox{%
    \vskip 0.1in
    \begin{center}
      \parbox{5.5in}{\centering\LARGE\bf \@title\par}
    \end{center}
    \vskip 0.1in
    \centering
    \def\And{\end{tabular}\hfil\linebreak[0]\hfil%
      \begin{tabular}[t]{c}\ignorespaces}
    \def\AND{\end{tabular}\hfil\linebreak[4]\hfil%
      \begin{tabular}[t]{c}\ignorespaces}
    \begin{tabular}[t]{c}\@author\end{tabular}%
    \vskip 0.3in \@minus 0.1in}}
    
\renewenvironment{abstract}
  {\begin{quote}\textbf{Abstract:}}
  {\par\vskip 1ex\end{quote}}

\providecommand{\keywords}[1]{%
  \begin{quote}\textbf{Keywords:} #1\end{quote}%
  \ifdefined\hypersetup\hypersetup{pdfkeywords={#1}}\fi}

\makeatother

\title{Remember what you did?: Learning Behavioral Memories for Partially Observable Object Manipulation}

\author{%
  \begin{tabular}{Y{1.2in} Y{1.2in} Y{1.2in}}
    {\large Kuancheng Wang} & {\large Seungho Yeom$^{*}$} & {\large Jinglin Cao$^{*}$} \\
    {\large Yuheng Zhi$^{\dagger}$} & {\large Nikhil Shinde$^{\dagger}$} & {\large Michael Yip} \\
  \end{tabular} \\[0.3em]
  UC San Diego \\
  $^{*}$Equal contribution \quad $^{\dagger}$Equal advising \\
  Website: \href{https://robo-camp.github.io/}{\textcolor{myorange}{\texttt{robo-camp.github.io}}}
}
\date{}

\begin{document}
\maketitle
\begin{figure}[H]
    \centering
    \includegraphics[width=1.0\textwidth]{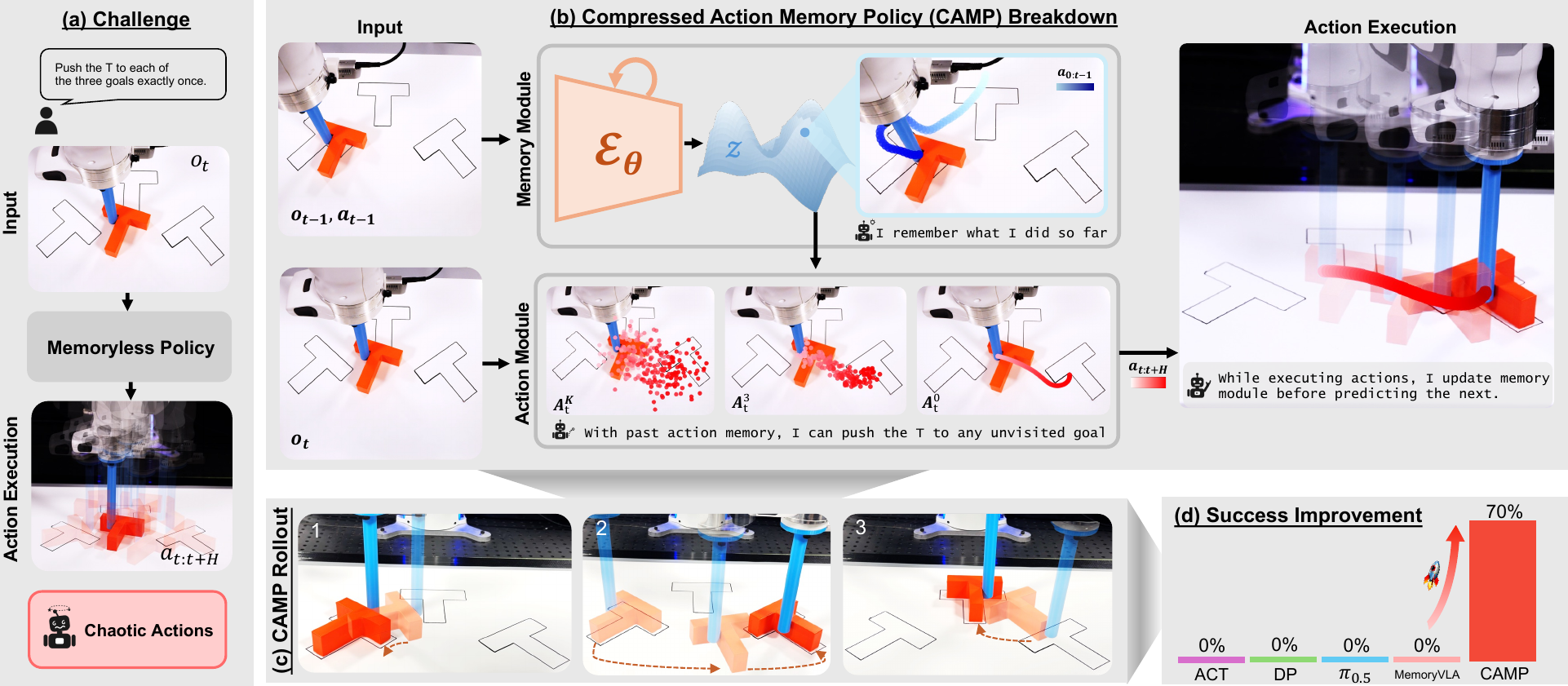}
    \caption{\textbf{From 0\% to 70\% success: a compressed memory of past actions resolves partial observability.} Long horizon, contact-rich manipulation tasks are partially observable, so memoryless policies act on an aliased signal and fail on tasks like Push-T-Multi-Goals (\emph{left}). CAMP (\emph{right}) pretrains a memory module $\mathcal{E}_\theta$ to reconstruct the past action trajectory, yielding a compressed code z that conditions a diffusion action head and turns the partially observed problem Markovian.}
    \label{fig:teaser}
\end{figure}

\begin{abstract}
Long horizon, contact-rich manipulation is inherently partially observable. 
This is as a single visual observation rarely captures a robot's full action context, including prior attempts, interactions, or progress. 
Consequently, standard visuomotor policies or vision-language-action models are prone to struggle in such tasks due to a lack of memory.
To address this, we introduce Compressed Action Memory Policy (\methodname) based on the insight that a robot’s own action history serves as a highly informative, self-supervised signal, enabling the policy to learn a robust, compact history representation. 
In our approach, we train a  memory module to maintain a compressed representation of past actions, forcing it to encode a latent behavioral memory of all the robot’s past interactions that can then be used to better contextualize future actions. 
This allows our approach to implicitly track generalized task progress and learn from failed attempts without any additional supervision, or external oversight. 
We evaluate \methodname across four real-robot setups and two novel simulation benchmarks: Memory-T-Bench and Memory-Manip-Bench. 
By demonstrating substantial gains over state-of-the-art baselines, \methodname is, to our knowledge, the first policy to demonstrate substantial success on contact-rich partially observable manipulation tasks purely through learned memory.
\end{abstract}


\section{Introduction}
To achieve true generalizable autonomy in unstructured real-world environments, robots must be capable of executing complex, extended objectives. 
Meaningful real-world applications, from household applications like cooking or cleaning in the kitchen, to industrial uses such as the safe assembly of tight tolerance components, are all inherently long-horizon and contact-rich. 
They all demand a sequence of complex physical interactions, the ability to safely recover from unexpected physical disturbances, and the seamless linking of multi-goal objectives. 

The inherent crux of this challenge comes from partial observability. 
Policies that rely solely on immediate and recent visual observations suffer from amnesia, as they fail to capture critical context, such as prior failed attempts, the phase of the current task, or even the history of the robot’s physical interactions. 
While standard diffusion policies falter in these settings even when augmented with basic observation history, existing memory-augmented alternatives also face severe limitations. 
Vision-Language-Action (VLA) based approaches often require users to carefully craft task-specific prompts to dictate what the robot should remember for such long horizon objectives. 
This reliance on human domain knowledge is tedious and unscalable. 
Forcing a model to learn history by predicting high-dimensional past or future visual observations quickly becomes computationally prohibitive and often exhausts model capacity on learning task-irrelevant visual details. 

To address these limitations and enable robots to solve extended objectives entirely autonomously we introduce the \textbf{\underline{C}}ompressed \textbf{\underline{A}}ction \textbf{\underline{M}}emory \textbf{\underline{P}}olicy (CAMP). 
Our core insight is intuitive, just as a human relies on the memory of what they just physically felt or attempted, a robot can leverage its own action history as a highly informative, self-supervised signal. 
By learning to predict its own past actions, the system can naturally learn to extract and retain only the most critical task-relevant behavioral information. 

Specifically, CAMP trains a recurrent 
memory module to predict a compressed representation - the Discrete Cosine Transform (DCT) -  of past action sequences. 
The resulting latent history representation successfully captures crucial behavioral context, such as the previous physical interactions and intermediate accomplished objectives. 
By providing this representation as an additional observation to our diffusion policy, CAMP effectively gains the memory required to execute complex, multi-step tasks in partially observable environments, bypassing the need for expensive visual prediction, manual human prompting and manual human task labelling.
We validate CAMP across a rigorous suite of environments demonstrating substantial gains over state-of-the-art visuomotor and VLA baselines. 
To our knowledge, CAMP is the first policy to resolve contact-rich, partially observable manipulation challenges purely through learned memory. 
In summary, our core contributions are: 
\begin{itemize}
\item \textbf{Self-Supervised Memory Formulation:} the introduction of action history reconstruction as a scalable, self-supervised signal to capture task-relevant behavioral memory 
\item \textbf{The  CAMP Architecture:} A novel framework integrating an recurrent memory module trained via DCT action compression with a diffusion policy to enable long horizon contact-rich manipulation tasks. 
\item \textbf{Novel Benchmarks:} The release of Memory-T-Bench and and Memory-Manip-Bench, two novel simulation benchmarks explicitly designed to evaluate memory-dependent, contact-rich tasks.
\item \textbf{Extensive Validation:} Through empirical evaluation demonstrating CAMP’s strength across these new simulation benchmarks and four complex real robot setups, yielding insights into memory representation.
\end{itemize}
\section{Related Work}
\paragraph{Memory-dependent robotic manipulation benchmarks.}
Recent benchmarks have begun to expose memory as a core bottleneck in long-horizon robotic manipulation. MemoryBench in SAM2Act~\cite{fang2025sam2actintegratingvisualfoundation}, RoboMME~\cite{dai2026robommebenchmarkingunderstandingmemory}, RMBench~\cite{chen2026rmbenchmemorydependentroboticmanipulation}, MIKASA-Robo~\cite{cherepanov2026memorybenchmarkrobots} and LIBERO-Mem~\cite{chung2025rethinkingprogressionmemorystate} study memory-dependent manipulation from spatial, temporal, object-centric, procedural, and reinforcement learning perspectives, showing that policies must often remember previous observations, interacted objects, repeated actions, or task progress. 
These benchmarks, however, are built almost entirely on quasi-static pick-and-place tasks, where the relevant memory is about what was seen or which object was handled. Contact-rich manipulation imposes a different and underexplored demand: the policy must remember what it has physically attempted -- prior contacts, the partial progress those contacts produced, and how earlier interactions changed the physical state -- since this history is often unobservable from any current frame. Our Memory-T-Bench and Memory-Manip-Bench target this action-interaction memory regime, isolating tasks where success hinges on remembering the robot's own past physical interactions.

\paragraph{Memory-augmented visuomotor policies.}
Existing memory-augmented policies mainly improve temporal context through visual-token memory, prompt memory, retrieval, or long-history encoding. SAM2Act+~\cite{fang2025sam2actintegratingvisualfoundation}, EchoVLA~\cite{lin2026echovlasynergisticdeclarativememory},  MemoryVLA~\cite{ shi2026memoryvlaperceptualcognitivememoryvisionlanguageaction}, CronusVLA~\cite{li2025cronusvlaefficientrobustmanipulation}, MAP-VLA\cite{ li2025mapvlamemoryaugmentedpromptingvisionlanguageaction}, MEM~\cite{torne2026memmultiscaleembodiedmemory}, and MemER~\cite{ sridhar2025memerscalingmemoryrobot} store or retrieve visual, semantic, or demonstration-derived memories for long-horizon VLA control; RoboFlamingo~\cite{li2024visionlanguagefoundationmodelseffective} uses LSTM~\cite{6795963} to remember recent action-relevant history and improve sequential decision-making; BPP~\cite{mark2026bpplongcontextrobotimitation}, and Past-Token Prediction~\cite{ torne2025learninglongcontextdiffusionpolicies} improve long-context conditioning through keyframe selection, full-history encoding, or auxiliary prediction; HiF-VLA~\cite{lin2026hifvlahindsightinsightforesight} and UniVLA~\cite{bu2025univlalearningacttaskcentric} explore compact motion or compress past action into prompts; and TraceVLA~\cite{zheng2025tracevlavisualtraceprompting} compressed the compresses the past actions into 2D traces and paints them on the observation with the difficulty to apply to long-term memory. These methods demonstrate the importance of memory, but they still largely rely on visual observations, language prompts, retrieved keyframes, or full observation histories. CAMP instead learns a compact behavioral memory directly from the robot's own action history, which is especially important for contact-rich tasks where progress is often encoded in physical attempts rather than visual appearance alone.

\section{Method}
\label{sec:method}
\begin{wrapfigure}{r}{0.4\textwidth}
\vspace{-1.cm}
    \centering
    \includegraphics[width=0.4\textwidth]{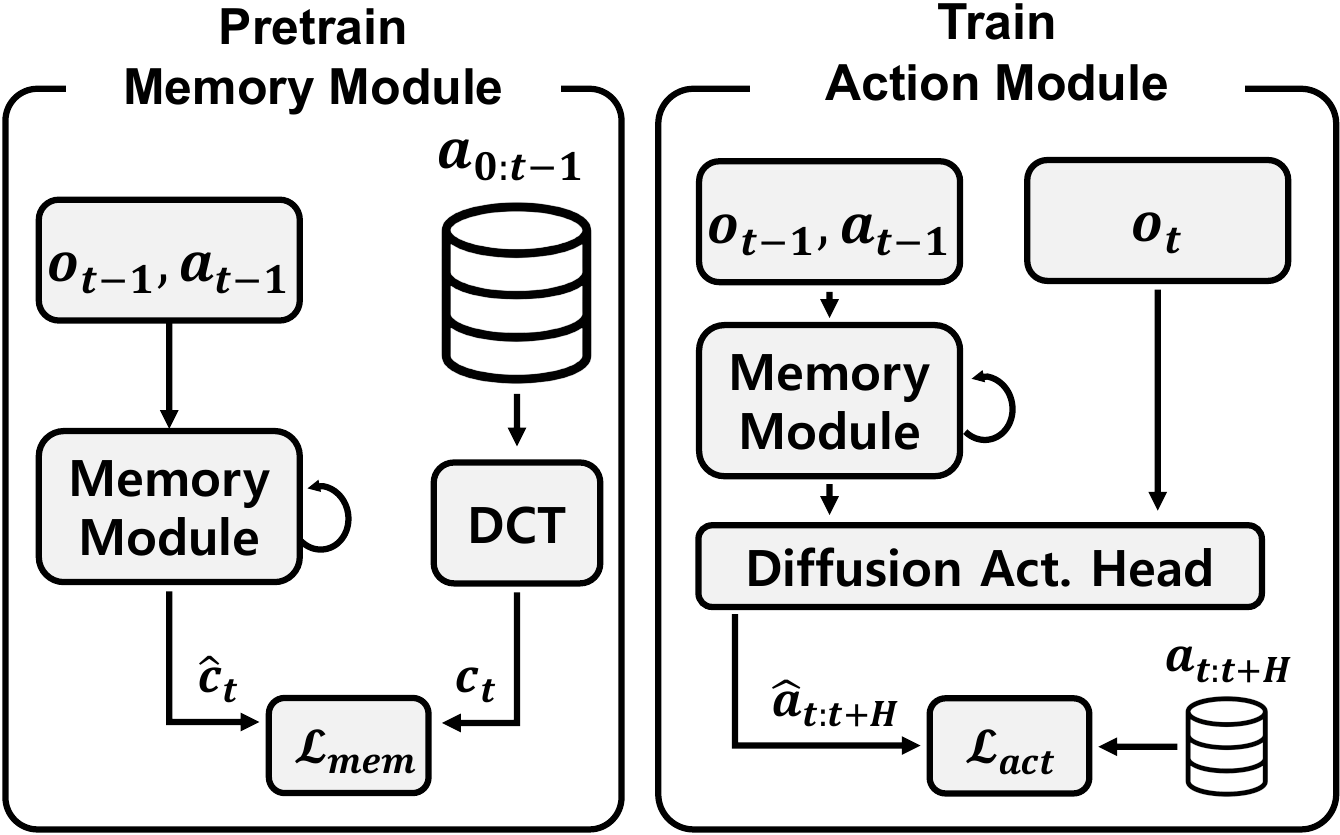}
    \captionsetup{font=small}
    \caption{The Training Pipeline of CAMP.}
    \label{fig:method_fig}
    \vspace{-0.6cm}
\end{wrapfigure}
We model vision-based manipulation as a partially observable Markov decision process (POMDP)
$(\mathcal{S}, \mathcal{A}, \mathcal{T}, \mathcal{O}, \mathcal{Z})$. At time $t$, the environment is
in an underlying state $s_t \in \mathcal{S}$. When the robot applies an action $a_t \in \mathcal{A}$, the state evolves
under a transition function $\mathcal{T}(s_{t+1} \mid s_t, a_t)$, and the robot receives only an observation $o_{t+1} \in \mathcal{O}$ drawn from an observation function $\mathcal{Z}(o_{t+1} \mid s_{t+1}, a_t)$.

In many cases, a single observation cannot, on its own, identify certain critical states of the task. An example of such states is task progress, which may be defined as sub-goals reached or objects already handled; the latest observation may not contain the full context to infer task progress. Thus, policy conditioned only on the current observation acts on an aliased signal and may be suboptimal. A standard remedy is to condition the next action on the full interaction history $\tau_{1:t} = (o_1, a_1, \dots, o_{t-1}, a_{t-1}, o_t)$. However, in vision-based manipulation tasks with high-dimensional observation spaces and high frequencies, processing the full history becomes impractical for real-time applications.

We consider how to summarize $\tau_{1:t}$ compactly yet sufficiently to contextualize the next action.
The Compressed Action Memory Policy (CAMP) builds on the observation that the past action component $a_{1:t-1}$ of $\tau_{1:t}$ is relatively lower dimensional and can be used as a self-supervised signal. 
In our setting, each observation consists of two RGB views and proprioception, $o_t = (o^{\mathrm{third}}_t, o^{\mathrm{wrist}}_t, p_t)$, and each action $a_t \in \mathbb{R}^{d_a}$ ($d_a = 7$) specifies an absolute end-effector target with a binary gripper command.
We use absolute rather than delta actions so that the past action sequence $a_{1:t-1}$ has consistent low-frequency structure, which is essential for the DCT-based memory objective introduced below.
We hypothesize that a memory module capable of reconstructing the robot's past decision trajectory $a_{1:t-1}$ has encoded almost all the information required to correctly generate the next action.
We train a recurrent memory module and a memory-augmented policy entirely from a dataset of expert demonstrations $\mathcal{D} = \{\tau^i\}$, respecting the assumptions of an imitation learning framework.
CAMP is built with: a recurrent memory module that summarizes the history trained under self-supervision (\Cref{sec:method:memory}) and action head
(\Cref{sec:method:vq,sec:method:stitch}). We close with the training schedule (Section~\ref{sec:method:schedules}). The training overview is provided in Figure~\ref{fig:method_fig}.

\subsection{Memory Module Pretraining}
\label{sec:method:memory}
We instantiate the recurrent memory module $\mathcal{E}_\theta$ as a two-layer LSTM~\cite{6795963} encoder that consumes the observation-action pairs one step at a time. 
At time $t$, its input is the previous action $a_{t-1}$ and observation $o^{\mathrm{third}}_{t-1}$ since, unlike the wrist camera frame, the third-view camera is spatially fixed, providing a consistent reference frame across time. We encode the image through a visual encoder $f_\psi$ (shared with the action head, Section~\ref{sec:method:stitch}) and concatenate the resulting feature with the action to form the per-step input 
$x_{t-1} = \texttt{concat}(f_\psi(o^{\mathrm{third}}_{t-1}),\, p_{t-1},\, a_{t-1})$.

The hidden state then updates as $h_t = \mathcal{E}_\theta(x_{t-1}, h_{t-1})$, where $h_t \in \mathbb{R}^{d_h}$ is a latent summary of the history so far, with $d_h$ the LSTM hidden-state dimension. The action head sees the past trajectory only through the memory module.

\label{sec:method:pretrain}
The pretraining objective predicts the robot's own past action trajectory $a_{t-L:t-1}$ from the hidden state $h_t$, where $L$ is the length of the past trajectory window (in our experiments, the full episode length per task), compressed in the frequency domain via the discrete cosine transform (DCT) so the memory keeps a behavioral summary rather than memorize the raw sequence. Actions are dominated by low-frequency structure, so a handful of low-order DCT coefficients capture the shape of the trajectory while discarding step-level detail.
Let $B \in \mathbb{R}^{L \times K}$ be the first $K$ orthonormal type-II DCT basis vectors of length $L$. A linear head $g_\phi$ maps the hidden state to coefficients $\hat{c}_t = g_\phi(h_t) \in \mathbb{R}^{K \times d_a}$, and the past trajectory is recovered as $\hat{a}^{\mathrm{past}}_t = B\hat{c}_t$. Since $K \ll L$, this reconstruction is lossy by construction, which is what makes $h_t$ a compact summary rather than a buffer. 
In practice, the action head predicts the full $d_a$-dimensional action, while memory
reconstruction targets only the action dimensions that vary on the task. The pretraining objective combines two terms.

\paragraph{Reconstruction loss.} We supervise in DCT coefficient space with a frequency weight that downweights high-frequency coefficients, forcing the memory to capture the low-frequency structure that dominates actions:
\begin{equation}
\mathcal{L}_{\mathrm{rec}} = \sum_{k=1}^{K} w_k \lVert \hat{c}_{t,k} - c_{t,k} \rVert_2^2, 
\qquad w_k = \exp\!\left(-\gamma\, \frac{k}{K-1}\right),
\label{eq:rec-loss}
\end{equation}
where $c_t = B^{\top} a_{t-L:t-1}$ are the coefficients of the demonstrated past trajectory, and $\gamma > 0$ controls how aggressively high-frequency coefficients are suppressed.

\paragraph{Temporal consistency.} A fixed past action is reconstructed from many timesteps as the episode advances, and these estimates should agree: slot $k$ of the reconstruction at time $t$ refers to the same action as slot $k+N$ at time $t+N$. We penalize disagreement between predictions $N$ steps apart,
\begin{equation}
\mathcal{L}_{\mathrm{cons}} = \sum_{N} \big\lVert \hat{a}^{\mathrm{past}}_{t}[\,{:}\,L{-}N\,] - \hat{a}^{\mathrm{past}}_{t+N}[\,N{:}\,] \big\rVert_2^2 ,
\label{eq:cons-loss}
\end{equation}
averaged over offsets $N$ and the valid overlap, so the memory's account of earlier events does not drift as new observations arrive.

The combined pretraining loss is
\begin{equation}
\mathcal{L}_{\mathrm{mem}} = \mathcal{L}_{\mathrm{rec}} + \lambda_{\mathrm{cons}}\, \mathcal{L}_{\mathrm{cons}},
\label{eq:pretrain-loss}
\end{equation}
where $\lambda_{\mathrm{cons}}$ weights the consistency term. We minimize $\mathcal{L}_{\mathrm{mem}}$ jointly over $\{\theta, \psi, \phi\}$ by backpropagation through time across each demonstration window, so that gradients from the reconstruction and consistency terms shape the recurrent dynamics of $\mathcal{E}_\theta$ in addition to the readout and visual encoder.

\subsection{Discretizing the Memory}
\label{sec:method:vq}
Supplying the continuous hidden states directly to the action head, letting it condition on the fine-grained trajectory history, hurts generalization in practice, as we will show in the experiment section. We therefore discretize $h_t$ with a vector quantizer that maps it to its nearest entry in a learned codebook $\mathcal{C} = \{e_1, \dots, e_{K_q}\}$ of size $K_q$:
\begin{equation}
z_t = \arg\min_{e \in \mathcal{C}} \lVert h_t - e \rVert_2,
\qquad
\tilde{h}_t = h_t + \mathrm{sg}[z_t - h_t],
\label{eq:vq-quantize}
\end{equation}
where $\mathrm{sg}[\cdot]$ denotes the stop-gradient operator and $\tilde{h}_t$ is the straight-through estimator that preserves gradient flow to the encoder. We train the codebook with the standard vector quantization (VQ) objective:
\begin{equation}
\mathcal{L}_{\mathrm{VQ}} = \lVert \mathrm{sg}[h_t] - z_t \rVert_2^2 + \beta \lVert h_t - \mathrm{sg}[z_t] \rVert_2^2,
\label{eq:vq-loss}
\end{equation}
where $\beta$ is the commitment weight. An MLP $g_\xi$ then projects the quantized state to the memory code 
$m_t = g_\xi(\tilde{h}_t) \in \mathbb{R}^{d_m}$, the memory's output dimension, 
which is the only signal the action head receives about the history.

\subsection{Diffusion Action Head}
\label{sec:method:stitch}
The action head follows Diffusion Policy~\cite{chi2024diffusionpolicyvisuomotorpolicy}, predicting an action chunk $a_{t:t+H}$ of horizon $H$ by iteratively denoising a Gaussian sample under a noise-prediction network $\epsilon_\eta(a^{k}, c_t, k)$ with parameters $\eta$, where $k$ indexes the diffusion step and $a^k$ is the partially-noised action chunk at diffusion step $k$. The memory enters only through the conditioning vector, which concatenates the shared visual encoder $f_\psi$ applied to both camera views, proprioception, and the memory code:
\begin{equation}
c_t = \texttt{concat}\bigl[f_\psi(o^{\mathrm{third}}_t),\; f_\psi(o^{\mathrm{wrist}}_t),\; p_t,\; m_t\bigr].
\label{eq:dp-cond}
\end{equation}
The wrist view supplies the close-range visual detail needed for precise contact-rich actions, while the memory module operates on the third-view stream alone, which is sufficient for tracking task progress. The head is trained with the standard score-matching loss
\begin{equation}
\mathcal{L}_{\mathrm{DP}} = \mathbb{E}_{a,\, k,\, \epsilon} \bigl\lVert \epsilon_\eta(a^{k}, c_t, k) - \epsilon \bigr\rVert_2^2.
\label{eq:dp-loss}
\end{equation}

\subsection{Training Schedules}
\label{sec:method:schedules}
After pretraining the memory on $\mathcal{L}_{\mathrm{mem}}$, we train the action head under $\mathcal{L}_{\mathrm{act}} = \mathcal{L}_{\mathrm{DP}} + \mathcal{L}_{\mathrm{VQ}}$ using a staged \emph{warm-up-then-finetune} schedule: the memory stays frozen for the first $N_{\mathrm{warm}}$ gradient steps (the warm-up duration), then is unfrozen and updated jointly with the action head at a learning rate scaled by $\alpha < 1$. We adopt the warm-up because joint training from the start drives the hidden state toward immediate action prediction and erodes the long-horizon information pretraining installs, an effect we observe directly later in the ablation studies.
\section{Experiments}
We evaluate \methodname following a series of six questions, each isolating one part of the central claim that the compressed action memory resolves partial observability for visuomotor manipulation policies: 
\begin{enumerate}
    \item Does action memory close the partial-observability gap on contact-rich tasks?
    \item How does \methodname compare against memoryless, large-scale, or memory-enhanced baseline models on vision-based 3D manipulation tasks? 
    \item Does \methodname's success rate scale with the increasing number of demonstrations?
    \item Which components of the recipe  are responsible for the performance gain?
    \item What are the characteristics of the action memory representation needed by the action policy?
    \item Does the recipe transfer to real robot tasks?
\end{enumerate}

\subsection{Simulation Experiments}
\subsubsection{Benchmarks and Baselines}
We evaluate on two simulated benchmarks. Memory-T-Bench is a controlled 2D contact-rich pushing suite built on PushT~\cite{chi2024diffusionpolicyvisuomotorpolicy}: dynamics is fixed across
variants and only the kind of the partial observability varies, isolating the effect of memory.
Memory-Manip-Bench extends the recipe to vision-based 3D manipulation. 
Across both benchmarks, tasks are designed to require memory for two purposes: \emph{tracking task progress} and 
\emph{learning from failure}.

\paragraph{Memory-T-Bench.}
\begin{wrapfigure}{r}{0.5\textwidth}
\vspace{-0.5cm}
    \centering
    \includegraphics[width=0.5\textwidth]{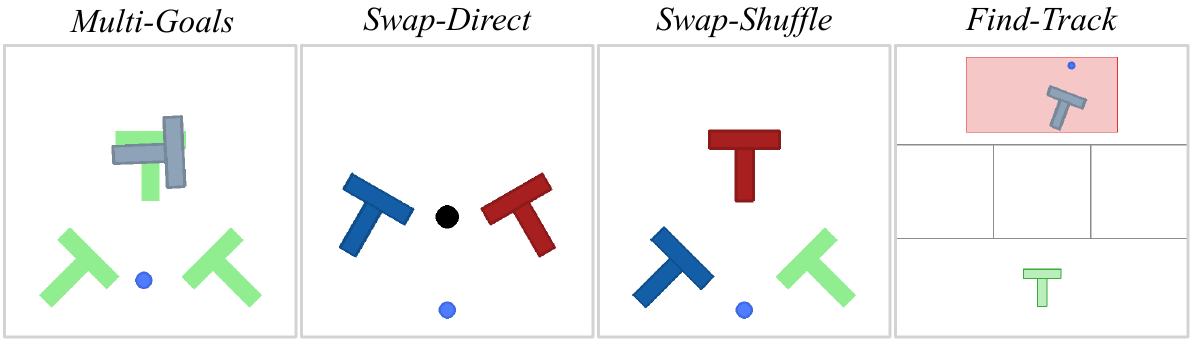}
    \caption{\textbf{Memory-T-Bench} tasks' initial states.}
    \label{fig:memory-t-bench}
\end{wrapfigure}
Four PushT-derived variants share the same contact-rich dynamics. The first three are \textit{track-task-progress} tasks: in \textit{Multi-Goals}, the agent pushes the T-block into three goal regions in any order without 
repetition; \textit{Swap-Direct} and \textit{Swap-Shuffle} require two T-blocks exchange positions, but mid-episode frames do not encode the original assignment, which Swap-Shuffle further randomizes across episodes. \textit{Find-Track}
is a learn-from-failure task: of the three pushing tracks, two have elevated friction that will block the T, and the robot must reach the goal by finding the low-friction track without trying the same track twice.

\paragraph{Memory-Manip-Bench.}
\begin{figure*}[tbp]
    \centering
    \resizebox{1.0\textwidth}{!}{\includegraphics{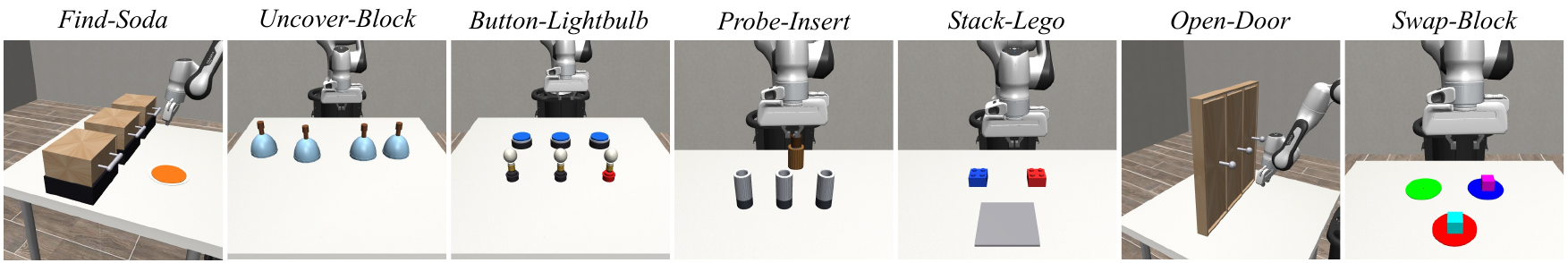}
    }
    \caption{\textbf{Memory-Manip-Bench} partially observable 3D manipulation tasks }
    \label{fig:memory_manip_bench
    }
\end{figure*}
We propose seven partially observable 3D tasks spanning the same two memory requirements and a range of contact.
Most are learn-from-failure: \textit{Find-Soda} opens closed drawers one at a time, skipping any
already found empty, until the soda is located; \textit{Uncover-Blocks} lifts each similar cover
once, moving on once it finds nothing underneath; \textit{Button-Lightbulb} probes buttons to infer
a hidden one-to-one button--bulb mapping before activating the lightbulb with a red base; \textit{Probe-Insert} probes
three holes until the one accepting the peg at full depth is found; \textit{Stack-Lego}
attempts a stack to discover which of two blocks has a stackable bottom; and \textit{Open-Door} adds short-horizon contact through the handle and hinge, where only one of the three doors is unlocked and can be pushed to open. \textit{Block-Swap} requires tracking the task progress, where it swaps two blocks via a buffer while recalling their initial positions.

\paragraph{Baselines.}
We compare against ACT~\cite{zhao2023learningfinegrainedbimanualmanipulation}, Diffusion
Policy~\cite{chi2024diffusionpolicyvisuomotorpolicy},
$\pi_{0.5}$~\cite{intelligence2025pi05visionlanguageactionmodelopenworld}, and
MemoryVLA~\cite{shi2026memoryvlaperceptualcognitivememoryvisionlanguageaction}. Diffusion Policy (DP) is
the memoryless counterpart sharing CAMP's action head, isolating memory; $\pi_{0.5}$ tests whether scale and broad pretraining substitute for explicit memory; MemoryVLA tests whether action memory improves on an existing memory mechanism. The VLA baselines run only on Memory-Manip-Bench, since their real-data pretraining makes the abstract 2D pushing domain out-of-distribution. The specific training details are provided in the Appendix~\ref{app:training-details}.

\subsubsection{Main Results}
\paragraph{The compressed action memory closes the partial-observability gap and imposes no penalty on fully observable tasks.}
\begin{table*}[t]
\centering
\small
\setlength{\tabcolsep}{0pt}
\captionsetup{skip=6pt}
\begin{tabular*}{\textwidth}{@{\extracolsep{\fill}}lccccc@{}}
\toprule
\textbf{Models} & \textbf{Push-T} & \textbf{Multi-Goals} & \textbf{Swap-Direct} & \textbf{Swap-Shuffle} & \textbf{Find-Track} \\
\midrule
DP & 94.0 & 56.0 & 74.0 & 48.0 & 62.0 \\
\textbf{CAMP (ours)} & \textbf{98.0} & \textbf{94.0} & \textbf{94.0} & \textbf{92.0} & \textbf{90.0} \\
\bottomrule
\end{tabular*}
\caption{Success rates in \% on the Memory-T-Bench, averaged over 50 trials per task.}
\label{tab:memory-t-bench}
\end{table*}
Table~\ref{tab:memory-t-bench} shows that CAMP improves over the memoryless DP on every
Memory-T-Bench variant, and the extent of the improvement tracks how much the observation hides. On
plain Push-T, which is Markovian, the two perform comparably (98.0\% vs.\ 94.0\%), showing that CAMP's memory imposes no penalty when the current frame already determines the action and memory is unnecessary. 
The gap widens sharply once the observation no longer identifies task progress or contains traces of past failed attempts,  94.0\% vs.\ 56.0\% on Multi-Goals, and 94.0\% vs.\
74.0\% and 92.0\% vs.\ 48.0\% on Swap-Direct and Swap-Shuffle, where the original placement of the blocks is not recoverable from any single frame. 
The fact that the margin grows precisely with the degree of partial observability, rather than uniformly across tasks, indicates that the gain comes from memory resolving hidden state.

\paragraph{The memory capability transfers to vision-based 3D manipulation.}
\newcolumntype{C}{>{\centering\arraybackslash}X}
\begin{table*}[t]
\centering
\small
\setlength{\tabcolsep}{3pt}
\begin{tabularx}{\textwidth}{@{}l *{4}{C}@{}}
\toprule
\textbf{Method} & \textbf{Avg. SR} & \textbf{Swap-Block} & \textbf{Button-Lightbulb} & \textbf{Uncover-Blocks} \\
\midrule
ACT    & 17.1          & 0.0         & 26.0         & 8.0          \\
DP   & 40.9          &  18.0        & 6.0          & 66.0          \\
Pi0.5& 26.8 & 26.0 & 14.0 & 68.0 \\
MemoryVLA & 5.7 & 0.0 & 0.0 & 6.0 \\
\textbf{CAMP (ours)} & \textbf{64.3} & \textbf{86.0} & \textbf{36.0} & \textbf{90.0} \\
\midrule
\textbf{Method} & \textbf{Stack-Lego} & \textbf{Probe-Insert} & \textbf{Find-Soda} & \textbf{Open-Door} \\
\midrule
ACT    & 40.0          & 42.0         & 4.0          & 0.0          \\
DP  & 54.0  & 40.0 & 48.0 & \textbf{54.0} \\
Pi0.5 & 20.0 & 10.0 & 8.0 & 42.0 \\
MemoryVLA & 12.0 & 0.0 & 0.0 & 22.0 \\
\textbf{CAMP (ours)} & \textbf{72.0} & \textbf{54.0} & \textbf{62.0} & 50.0 \\
\bottomrule
\end{tabularx}
\caption{Success rates (SR) in \% on the Memory-Manip-Bench, averaged over 50 trials per task.}
\label{tab:memory-manip-bench}
\end{table*}

Table~\ref{tab:memory-manip-bench} extends the comparison to seven 3D manipulation tasks. CAMP reaches the highest average success rate of 64.3\%, more than 23\% above the strongest baseline (DP at 40.9\%), and achieves the best per-task result on six of the seven tasks. The margin is largest where the memory function is most demanding: 86.0\% vs.\ 18.0\% on Swap-Block,
which requires recalling initial state. CAMP also leads on the five learn-from-failure tasks (e.g., \textit{Uncover-Blocks}, \textit{Probe-Insert}) by up to 24.0\% over DP. The one task where CAMP does not lead is Open-Door, on which it trails DP by 4.0\%; we identified Open-Door as a short-horizon contact task with no clear memory demand when there are only three doors, and the result is consistent with the broader pattern that CAMP's advantage tracks partial observability rather than appearing uniformly across tasks. We note that MemoryVLA, the most directly comparable prior memory-based method, averages only 5.7\% on this suite, but attributes most of this gap to its observation modality
rather than its memory mechanism: MemoryVLA conditions on a single third-person RGB view and a language
instruction without the wrist-camera view or any depth information, which limits accurate grasping of small features
such as cover handles and small blocks. We therefore cannot directly compare its memory capability with our method.

\paragraph{\methodname converts more demonstrations into higher success; the memoryless policy does not.}
\begin{wrapfigure}{r}{0.45\textwidth}
\vspace{-0.8cm}
    \centering
    \includegraphics[width=0.45\textwidth]{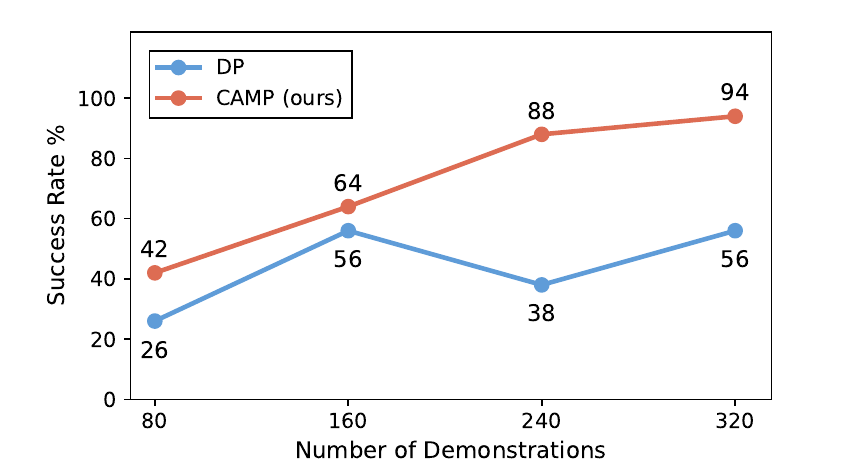}
    \caption{Success rate of DP and \methodname with increasing number of demos.}
    \label{fig:success_rate_vs_demo}
\end{wrapfigure}
Figure~\ref{fig:success_rate_vs_demo} varies the number of demonstrations on the \textit{Multi-Goals} task in the Memory-T-Bench. CAMP improves
monotonically with data, from 42\% at 80 demonstrations to 94\% at 320 demonstrations, whereas the memoryless
diffusion policy shows no consistent gain over the same range, fluctuating between 26\% and 56\%
without trending upward. The contrast indicates that the bottleneck for the memoryless policy is
not data but partial observability: additional demonstrations cannot teach a single-frame policy to
disambiguate states that look identical in the current observation, so its success rate stalls,
while CAMP's memory lets it turn the same additional data into steadily better performance. 

\paragraph{Contribution of each training recipe component.}
Table~\ref{tab:training_strategy_ablation} ablates the three design decisions behind CAMP, and each
is necessary to reach the full 94\% success rate. Removing the vector quantizer drops performance
to 60\%: without discretization the action head conditions on the continuous hidden state and keys on trajectory identity rather than the coarser behavioral state, so the memory generalizes poorly across demonstrations. The reconstruction target matters next: replacing the DCT basis with a Fourier (DFT) or raw-waypoint target drops performance to 64\% and 66\%, indicating that compressing the trajectory into a few low-frequency DCT coefficients\, rather than reconstructing it in a basis that spreads information across many components, is what forces the memory into a compact behavioral summary. The training schedule is the most delicate decision. Jointly training the memory and action head from scratch reaches only 66\%, since the action loss drives the hidden state
toward immediate prediction and never installs long-horizon memory. Pretraining the memory module and then freezing it when stitched to DP does better at 80\%, confirming that the pretrained representation is already informative, but it leaves performance on the table by never adapting the memory to the policy. \begin{wraptable}{r}{0.5\textwidth}
  \vspace{0.35\baselineskip}  
  \centering
  \footnotesize
  \setlength{\tabcolsep}{4pt}
  \renewcommand{\arraystretch}{0.8}
  \caption{Ablating each component of the training recipe on Multi-Goals in Memory-T-Bench.}
  \label{tab:training_strategy_ablation}
  \begin{tabular}{@{}ll@{\hspace{6pt}}c@{}}
    \toprule
    Component & Variant & Success \% \\
    \midrule
    Recon.\ target   & Raw waypoints       & 66 \\
                     & DFT                 & 64 \\
    \midrule
    Discretization   & No VQ               & 60 \\
    \midrule
    Training         & Joint (scratch)     & 66 \\
    schedule         & Pretrain + freeze   & 80 \\
                     & Pretrain + finetune & 46 \\
    \midrule
    \textbf{Full (CAMP)} & \textbf{VQ + DCT + staged} & \textbf{94} \\
    \bottomrule
  \end{tabular}
  \vspace{-0.65\baselineskip}
\end{wraptable} Pretraining the memory module and then immediately finetuning with DP's action loss is worst of all at 46\%, as joint gradients erase the pretrained memory before it can be exploited. 
Only our \emph{warm-up-then-finetune} schedule -- let the action head adapt to the frozen memory, then finetune jointly -- can recover the full 94\% success by preserving the pretrained memory and adapting it to the action head.

\paragraph{The best memory for the policy is not the best reconstructor.}
A natural assumption is that the best memory for the action head is the one that most faithfully encodes the full past action trajectory. Our results suggest the \textit{opposite}. Before finetuning, the memory module remains the accurate reconstructor produced by pretraining. With our \emph{warm-up-then-finetune} schedule, in which gradients from the action loss are allowed to flow back into the memory module, the finetuned memory reaches 94\% success, in contrast to the 80\% success before finetuning. Yet when we decode the past action trajectory from the finetuned hidden state, the reconstruction
is degraded relative to the pretrained one:
the reconstruction loss $\mathcal{L}_{\mathrm{rec}}$ rises from 0.35 before finetuning to 1.46
after. The finetuned memory has moved away from preserving the trajectory in detail, even though it has become more useful to the policy.

This pattern suggests that the policy does not need the memory to \textit{perfectly} recover the history, but it needs the memory to surface a compressed summary aligned with the decision the action head has to make.
Pretraining installs the long-horizon information the observation alone cannot provide, but finetuning shapes that information into a representation matched to the action distribution, at the cost of fidelity to the original reconstruction objective. The reconstruction loss is therefore best read as a \emph{scaffold} that forces the memory to absorb history during pretraining, rather than as the objective the policy ultimately benefits from. A useful action memory representation is one that retains sufficient behavioral structure to disambiguate states but discards detail that would otherwise compete with the action loss for capacity in the hidden state or hurt generalization.

\subsection{Real-Robot Experiments}
\begin{figure*}[tbp]
    \centering
    \resizebox{1.0\textwidth}{!}{\includegraphics{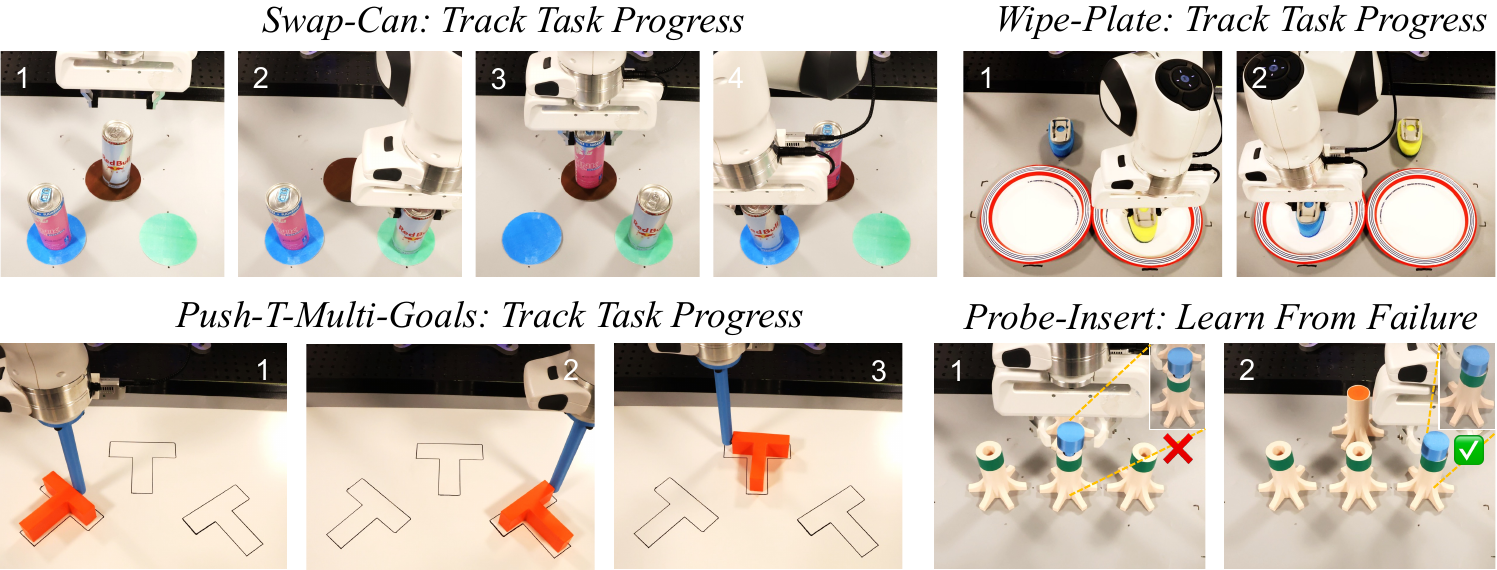}

    }
    \caption{\textbf{Real-robot tasks} visualized in multi-stage or with potential failed attempts.}
    \label{fig:real_world_task}
\end{figure*}
\subsubsection{Tasks and Baselines}
We evaluate \methodname on four real-robot tasks, as illustrated in Figure~\ref{fig:real_world_task} with the memory functions identified in simulation, \textit{tracking task progress} and \textit{learning from
failure}.
\textit{Push-T-Multi-Goals}, \textit{Swap-Can}, and \textit{Probe-Insert} follow the same protocols as \textit{Multi-Goals} in Memory-T-Bench and \textit{Block-Swap} and \textit{Probe-Insert} in Memory-Manip-Bench. \textit{Wipe-Plate} presents two identical plates and two brushes: the robot wipes one plate with a random brush, returns it to its start position, then must wipe the \emph{other} plate with the \emph{other} brush. Because the brushes return to their original positions and the plates are identical, the second wipe is undetermined from the current image alone.  We collect 100 demonstrations per task by teleoperating Franka Emika Panda robot through a Phantom Omni and train the same baselines in Memory-Manip-Bench. The set-ups are provided in Appendix~\ref{app:real-robot-setup}.
\subsubsection{Main Results}
\paragraph{Real-world evaluation exposes the memory bottleneck more starkly than simulation.}
On the real robot, every memoryless baseline collapses to 0/10 on Push-T-Multi-Goals, Swap-Can, and Wipe-Plate, while \methodname{} reaches 7/10 on each (Table~\ref{tab:real-world-table}). The failures are not uniform, and watching the rollouts clarifies what each baseline can and cannot do.

On \emph{Push-T-Multi-Goals}, every baseline can push the T-block roughly toward one of the goal regions, but none aligns it: without memory of which goal it is committing to, the policy oscillates between goal candidates and never settles into a final placement. Although this task requires low precision, MemoryVLA is still unable to complete. 

On \emph{Swap-Can}, baselines can complete at most the first two sub-steps -- placing one can on the green buffer and the second can on the first can's original location -- after which the policy stalls, with no memory of which can has already been moved and where the swap should terminate.

On \emph{Wipe-Plate}, baselines exhibit several distinct failure modes that all trace to the same memory deficit: reusing the same brush on the same plate, switching brushes but returning to a plate that has already been wiped, or keeping the first brush and wiping the second plate with it. $\pi_{0.5}$ in particular grasps the first brush reliably but then enters an infinite loop wiping a single plate, never switching to the second. MemoryVLA additionally suffers from imprecise grasping, which compounds with its failure to track which brush–plate pair remains.

On \emph{Probe-Insert}, DP and ACT achieve non-zero success rates only because of their first guess at the correct slot, in which case the peg is fully inserted on that attempt; on the remaining trials they cannot recover, since they have no memory of which slots have already been ruled out. MemoryVLA's bottleneck here is insertion precision.

\methodname{} is the only method that consistently completes all four tasks. We provide rollout videos of all baselines and \methodname{} on our project website for a more intuitive view of these failure modes.
\begin{table*}[t]
\centering
\small
\captionsetup{skip=3pt}
\begin{tabular*}{\textwidth}{@{\extracolsep{\fill}}lcccc@{}}
\toprule
\textbf{Models} & \textbf{Push-T-Multi-Goals} & \textbf{Probe-Insert} & \textbf{Swap-Can} & \textbf{Wipe-Plate} \\
\midrule
ACT  & 0/10  & 4/15 & 0/10 & 0/10 \\
DP & 0/10  & 5/15 & 0/10 & 0/10 \\
Pi0.5 & 0/10 & 0/15 & 0/10 & 0/10 \\
MemoryVLA & 0/10& 0/15 & 0/10 & 0/10 \\
\textbf{CAMP (ours)} & \textbf{7/10} & \textbf{11/15} & \textbf{7/10} & \textbf{7/10} \\
\bottomrule
\end{tabular*}
\caption{The success rate on four tasks reported as successes per trial. \methodname is the only method that completes any of these tasks at non-trivial rates.}
\label{tab:real-world-table}
\end{table*}
\section{Limitations and Conclusion}
\label{limitation_and_conclusion}
We introduced CAMP, a memory-augmented visuomotor policy for long-horizon, contact-rich manipulation under partial observability that turns the robot's own action history into a scalable, self-supervised learning signal. CAMP demonstrates consistent gains across Memory-T-Bench, Memory-Manip-Bench, and multiple challenging real-robot tasks.
Limitations remain. Tasks requiring extremely-long-horizon memory are still challenging for our method. Additionally, extending CAMP to dexterous manipulation is a natural direction we leave open for future work.

\clearpage
\bibliography{references}
\clearpage
\appendix
\section{Training Details}
\label{app:training-details}
\subsection{Visuomotor Policies}
The \methodname{}, Diffusion Policy (DP), and ACT baselines are trained from scratch independently per task on a single H100 GPU, with the same recipe applied across simulation and real-world tasks. For \methodname{}, we first pretrain the LSTM~\cite{6795963} memory module to reconstruct the past action trajectory in DCT coefficient space, supervising over the full episode length $L$ of each task. We then train the action head under the warm-up-then-finetune schedule, scaling the LSTM learning rate by $\alpha = 0.1$ during finetuning. Both stages use AdamW with EMA on the policy weights and a DDPM~\cite{ho2020denoisingdiffusionprobabilisticmodels} training scheduler combined with DDIM~\cite{song2022denoisingdiffusionimplicitmodels} inference. Full hyperparameter
values are listed in Table~\ref{tab:camp-hyperparams}. For DP, we follow the original recipe~\cite{chi2024diffusionpolicyvisuomotorpolicy}, with identical optimizer, batch size, EMA, and diffusion-scheduler settings, omitting the memory module. For ACT, we use a transformer encoder--decoder action head with chunk size $H = 100$, KL weight $10$, AdamW~\cite{loshchilov2019decoupledweightdecayregularization} (lr=$1\times 10^{-5}$, weight decay
$1\times 10^{-4}$), and batch size $256$.

\begin{table}[h]
\centering
\small
\setlength{\tabcolsep}{6pt}
\renewcommand{\arraystretch}{1.1}
\captionsetup{skip=4pt}
\caption{\textbf{\methodname{} Hyperparameters.}}
\label{tab:camp-hyperparams}
\begin{tabular}{@{}lc@{}}
\toprule
\textbf{Hyperparameter} & \textbf{Value} \\
\midrule
\multicolumn{2}{l}{\emph{Memory Module (LSTM) Pretraining}} \\
\quad Hidden dimension $d_h$              & 64 \\
\quad DCT coefficients $K$                & 32 \\
\quad Memory code dimension $d_m$         & 32 \\
\quad Batch size                          & 16 \\
\midrule
\multicolumn{2}{l}{\emph{Vector Quantizer}} \\
\quad Codebook size $K_q$                 & 128 \\
\quad Commitment weight $\beta$           & 0.25 \\
\midrule
\multicolumn{2}{l}{\emph{Diffusion Action Head Staged Training}} \\
\quad Action chunk size $H$               & 8 \\
\quad Diffusion training steps (DDPM)     & 100 \\
\quad Inference steps (DDIM)              & 16 \\
\quad EMA decay                           & 0.75 \\
\quad Batch size                          & 256 \\
\midrule
\multicolumn{2}{l}{\emph{Optimization}} \\
\quad Optimizer                           & AdamW \\
\quad Learning rate                       & $1\times 10^{-4}$ \\
\quad LSTM learning-rate scale $\alpha$   & 0.1 \\
\quad Weight decay                        & $1\times 10^{-6}$ \\
\quad LR schedule                         & Cosine \\
\bottomrule
\end{tabular}
\end{table}

\subsection{VLA Baselines}
Both $\pi_{0.5}$ and MemoryVLA baselines are fine-tuned independently per task using Low-Rank Adaptation (LoRA) on 4 H100 GPUs, with the same LoRA recipe applied across both the simulation and real-world tasks.

For $\pi_{0.5}$, we initialize from the $\pi_{0.5}$-base checkpoint and apply LoRA with rank $r=32, \alpha=64$ to the Gemma expert attention projections and rank $r=24, \alpha=48$ to the PaliGemma language model layers, with action projection layers trained in full. We use AdamW~\cite{loshchilov2019decoupledweightdecayregularization}  (lr=$2.5\times10^{-4}$, cosine schedule, weight decay 0.01) with a global batch size of 512 under BF16 mixed precision.

For MemoryVLA, we use two different base checkpoints depending on the task domain: the libero-100 checkpoint for Memory-Manip-Bench tasks and the memvla-bridge checkpoint for real-world tasks, reflecting the respective pre-training data distributions. As the action space dimensionality differs between domains, we apply per-domain DOF conversion to the demonstration data prior to training to align with each checkpoint's action head. In both cases, LoRA is applied at rank r=32 to the DiT~\cite{peebles2023scalablediffusionmodelstransformers} attention layers, rank $r=24$ to the LLaMA backbone, and rank r=8 to the vision encoder. Training uses AdamW~\cite{loshchilov2019decoupledweightdecayregularization} (lr=$2.5\times10^{-4}$, constant schedule) with a global batch size of 256 under BF16 mixed precision with FSDP full-shard and image augmentation.

\section{Real Robot Setup}
\label{app:real-robot-setup}
\begin{wrapfigure}{r}{0.5\textwidth}
    \centering
    \vspace{-1em}
    \includegraphics[width=0.5\textwidth]{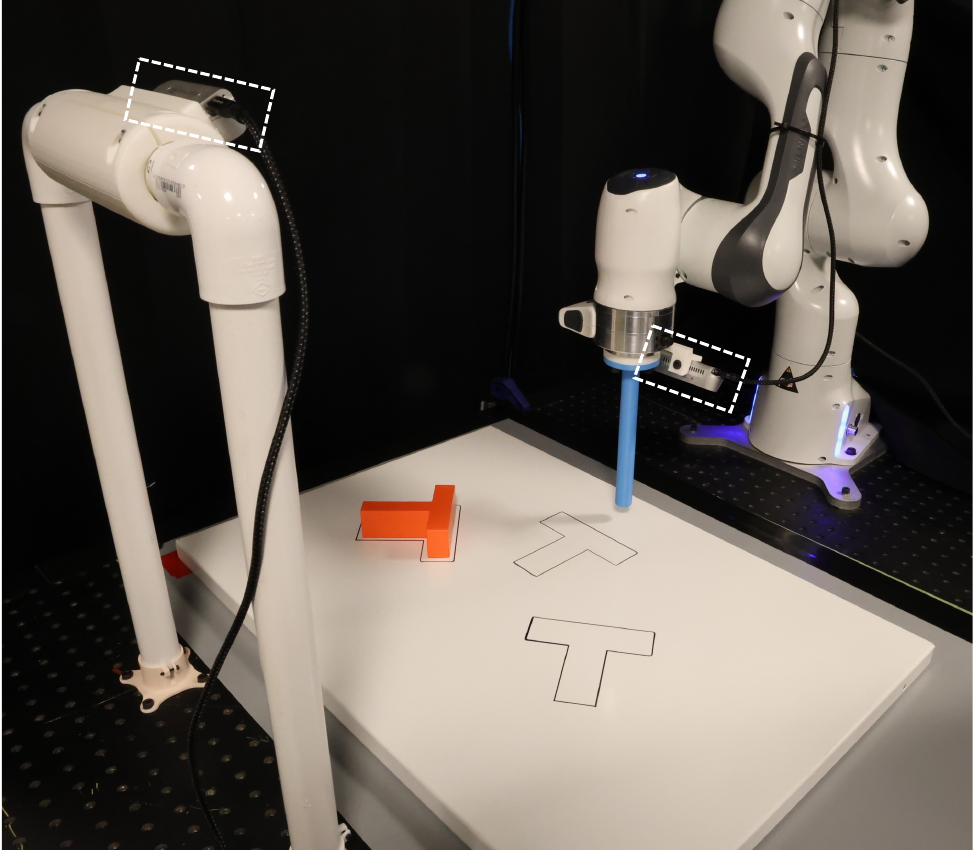}
    \caption{Real-robot setup. The white bounding boxes show the real-world camera set-up.}
    \label{fig:real-robot-set-up}
    \vspace{-1em}
\end{wrapfigure}
We use a Franka Emika Panda 7-DoF arm with two Intel RealSense D435 cameras: a third-person view mounted on a fixed stand and a wrist-mounted view rigidly attached to the end-effector (Fig.~\ref{fig:real-robot-set-up}). Demonstrations are collected via a Phantom Omni device for teleoperation, with 100 demonstrations recorded per task. At each timestep, we log both RGB streams together with the robot's proprioceptive state, which consists of the end-effector pose and the gripper state (open/closed). The action is an 7-dimensional vector comprising the 3-D end-effector position, a 3-D Euler orientation, and a 1-D gripper command. Consistent with the Memory-T-Bench tasks, the policy input consists of the two RGB images and the proprioceptive state, and the output is the robot action.
\end{document}